\documentclass{article}

\PassOptionsToPackage{numbers, compress}{natbib}

\usepackage[final]{neurips_2022}

\usepackage[utf8]{inputenc} 
\usepackage[T1]{fontenc}    
\usepackage{hyperref}       
\usepackage{url}            
\usepackage{booktabs}       
\usepackage{amsfonts}       
\usepackage{nicefrac}       
\usepackage{microtype}      
\usepackage{xcolor}         
\usepackage{tikz}
\usepackage{amsmath}
\usepackage{subcaption}

\usetikzlibrary{positioning}

\title{SurGNN: Explainable visual scene understanding and assessment of surgical skill using graph neural networks}

\author{%
  Shuja Khalid\thanks{Corresponding author} \\
  Department of Computer Science\\
  University of Toronto\\
  Toronto, ON M5S \\
  \texttt{skhalid@cs.toronto.edu} \\
  \And
  Frank Rudzicz \\
  Faculty of Computer Science \\
  Dalhousie University \\
  Halifax, NS B3H 4R2 \\
  \texttt{frank@dal.ca} \\
}

\begin{document}

\maketitle

\begin{abstract}
This paper explores how graph neural networks (GNNs) can be used to enhance visual scene understanding and surgical skill assessment. By using GNNs to analyze the complex visual data of surgical procedures represented as graph structures, relevant features can be extracted and surgical skill can be predicted. Additionally, GNNs provide interpretable results, revealing the specific actions, instruments, or anatomical structures that contribute to the predicted skill metrics. This can be highly beneficial for surgical educators and trainees, as it provides valuable insights into the factors that contribute to successful surgical performance and outcomes. SurGNN proposes two concurrent approaches -- one supervised and the other self-supervised. The paper also briefly discusses other automated surgical skill evaluation techniques and highlights the limitations of hand-crafted features in capturing the intricacies of surgical expertise. We use the proposed methods to achieve state-of-the-art results on EndoVis19, and custom datasets. The working implementation of the code can be found at https://github.com/<redacted>.
\end{abstract}



\section{Introduction}
Explainable visual scene understanding for surgical skill assessment using graph neural networks (GNNs) involves developing models that can interpret and reason about the complex visual data generated during surgical procedures. By representing the procedure as a graph structure and training a GNN on this data, we can model the dependencies and relationships between different surgical actions, instruments, and anatomical structures \cite{valderrama2022towards}. This allows us to extract relevant features from the visual data and predict various aspects of surgical skill, such as the overall quality of the procedure or the proficiency of individual actions \cite{seenivasan2022biomimetic}. This can help surgical educators and trainees better understand the underlying factors that contribute to successful surgical outcomes and improve training and performance \cite{maertens2016systematic, evans2016evolving, parsons2011surgical}.

Traditional assessment methods often rely on subjective expert ratings or global performance metrics \cite{gray1996global, martin1997objective, hogle2014evaluation} which can be biased, inconsistent, or difficult to interpret \cite{hogle2014evaluation, yan2014evaluation, samargandi2017objective, jayaraman2009objective}. By contrast, GNN-based models can provide a more objective and fine-grained analysis of surgical performance by interpreting the rich visual data available from surgical videos \cite{valderrama2022towards, seenivasan2022biomimetic}. GNN-based models can also provide explanations for their predictions, which can help build trust in the models and facilitate learning and improvement \cite{ying2020understanding, ji2019graph, kipf2017semi}. Additionally, we use 3D representations in our analysis. To the best of our knowledge, we are the first paper to consider explainable 3D representations for surgical skill analysis directly from visual cues.

\section{Related work}
\label{sec:related-works}
\paragraph{Automated surgical skill evaluation}
Automated surgical skill evaluation is an active research area, with various approaches proposed for analyzing surgical data using computer vision and machine learning \cite{zia2018automated, fard2018automated, levin2019automated, khalid2022or, chadebecq2023artificial}. A popular approach is to use hand-crafted features, such as motion analysis or tool usage, to train classifiers for surgical skill assessment. For example, Gao {\em et al} \cite{gao2014jhu} developed a feature-based model that used motion analysis to classify suturing and knot-tying skill in surgical videos. However, these hand-crafted features are often limited by their ability to capture the complex and subtle movements that characterize surgical expertise.

To address these limitations, researchers have also explored deep learning techniques, such as convolutional neural networks (CNNs) and recurrent neural networks (RNNs), for automated surgical skill evaluation. For example, Cheng {\em et al} \cite{cheng2022artificial} developed a CNN-based model that used visual cues to assess the proficiency of laparoscopic suturing and knot-tying in a simulator. Similarly, Zia {\em et al} \cite{zia2018surgical} proposed an RNN-based model that used motion analysis to classify surgical skill in a simulation environment. However, these deep learning models often lack interpretability, making it difficult to understand the underlying factors that contribute to their predictions. Anastasiou {\em et al} \cite{anastasiou2023keep} described an action-aware Transformer with multi-head attention producing inter-video contrastive features to regress the manually-assigned skill scores. This approach, although novel, requires consensus on what is considered good performance and has only been tested in the JIGSAWS \cite{gao2014jhu} dataset, which consists of very short, simulated procedures.

To address this interpretability issue, recent research focuses on explainable AI, including attention mechanisms and GNNs, for automated surgical skill evaluation. For example, Hira {\em et al} \cite{hira2022video} developed an attention-based model that used a hierarchical CNN to identify relevant visual features for surgical skill assessment. Meanwhile, Ban {\em et al} \cite{ban2022concept} proposed a GNN-based model that represented surgical actions as nodes in a graph structure, allowing for a more fine-grained analysis of surgical skill. These approaches represent promising directions towards automated surgical skill evaluation systems that are both accurate and interpretable.

\paragraph{Scene understanding using graph neural networks}
Scene understanding using GNNs is an active research area, with numerous approaches for leveraging the rich visual data available in complex scenes \cite{liang2021visual, fan4238333interpretable}. One popular application of GNNs in scene understanding is semantic segmentation, where the goal is to assign a semantic label to each pixel in an image \cite{wang2017gated}. For example, Wang {\em et al} \cite{wang2017gated} proposed a GNN-based model that used graph convolutions to incorporate contextual information from neighboring pixels, improving the accuracy of semantic segmentation in complex scenes.

Another popular application of GNNs in scene understanding is object detection and recognition, where the goal is to identify and localize objects in an image. Li {\em et al} \cite{li2022crowd} developed a GNN-based model that used spatial and semantic information to perform object detection and semantic segmentation simultaneously. Qi {\em et al} \cite{qi2021offboard} proposed a GNN-based model that used message-passing to refine object proposals and improve object detection performance.

GNNs have also been used for more complex tasks in scene understanding, such as 3D object detection and reconstruction. For example, Zhang {\em et al} \cite{zhang2021pc} developed a GNN-based model that used graph convolutions to reason about the 3D geometry of objects and their relationships in a scene. Similarly, Yin {\em et al} \cite{yin2020lidar} proposed a GNN-based model that used message passing to refine 3D object proposals and improve 3D object detection performance.

Overall, these works demonstrate the versatility and effectiveness of GNNs in scene understanding, and highlight their potential for improving the performance of computer vision systems in a wide range of applications, including surgical skill evaluation.

\section{SurGNN: The model}

Our architecture consists of two self-supervised graph attention layers followed by a global mean pooling layer and, finally, a linear classification layer for training the model in a supervised manner with the available labels. To generate the graph embeddings, we consider the following approaches: \textbf{2D} We follow the feature extraction approach, presented in \cite{khalid2022or} where each instrument is segmented and tracked over the length of a clip. A set of features are calculated from this temporal data to generate the node embeddings. \textbf{3D} Traditional approaches, discussed in section \ref{sec:related-works} of this paper use either 2D representations, or robotic kinematic data for 3D analysis. Since kinematic ground truth data isn't available for laparoscopic videos, we use neural radiance fields \cite{gao2021dynamicnerf, pumarola2021dnerf, li2021neuralnsff, park2021hypernerf, khalid2022wildnerf, khalid2023refinerf} to generate dynamic scene renderings. The neural scene renderings are advantageous as they allow for fixed camera renderings. By fixing the position of the camera, we aim to remove the relative motion of the camera with respect to the scene. Using the same approach presented in \cite{khalid2022or}, we generate 3D representations without the additional noise associated with extraneous camera motion.

\begin{figure}[ht]
    \centering
    \includegraphics[width=0.8\textwidth]{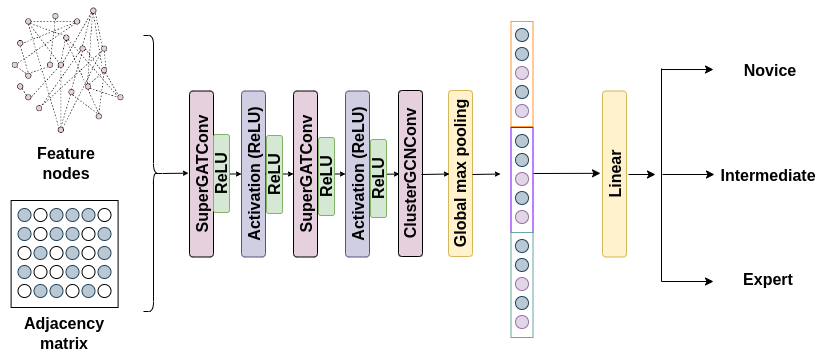}
    \caption{Architecture of the SurGNN model. Feature nodes are extracted in the pre-processing stage and the adjacency matrix generated. The resulting model is trained with a supervised objective using the provided labels. The self-supervised approach uses a \textit{spectral} objective.}
    \label{fig:arch}
\end{figure}

\subsection{Self-supervised Architecture}
The self-supervised architecture is inspired by  Tsitsulin \textit{et al.} \cite{tsitsulin2020graph}, who used a spectral loss, which is commonly used in GNNs to optimize the spectral properties of the learned representation. The spectral loss
is defined as the Frobenius norm of the difference between the spectral embeddings of the original graph and the reconstructed graph, and is typically used in conjunction with other loss functions,
such as cross-entropy loss. This is represented mathematically using the following equations:

The normalized Laplacian matrix L is calculated as follows:
\begin{equation}
    L = I - D^{-1/2} A D^{-1/2}
\end{equation}

We compute the eigenvectors $\{v_1, v_2, \ldots, v_n\}$ and eigenvalues $\{\lambda_1, \lambda_2, \ldots, \lambda_n\}$ of the Laplacian matrix $L$.

\begin{equation}
    L v_i = \lambda_i v_i, \quad i = 1, 2, \ldots, n
\end{equation}

For the self-supervised task, we randomly mask some nodes or edges in the graph. The goal is to predict the missing parts of the graph from its spectral representation:

\begin{equation}
    \text{Original Laplacian Matrix:} \quad L_{\text{original}} = L
\end{equation}
\begin{equation}
    \text{Reconstructed Laplacian Matrix:} \quad L_{\text{reconstructed}} = \text{GNN\_Decoder}(v_1, v_2, \ldots, v_n)
\end{equation}

The spectral loss can be defined using the mean squared error (MSE) between the original and reconstructed Laplacian matrices.

\begin{equation}
    \text{Spectral Loss} = \frac{1}{n} \sum_{i=1}^{n} \| L_{\text{original}}^{(i)} - L_{\text{reconstructed}}^{(i)} \|_F^2
\label{eq:spectral}
\end{equation}

The first part of the model architecture in Figure \ref{fig:arch} serves as the encoder which takes the graph representation of the input and creates an embedding of length 32. Masking out each embedding ans using this  loss function for training encourages the embeddings to capture the spectral properties of the graph, such as smoothness and locality, while also preserving the modularity structure of the network. By using this loss, we are able to systematically train the model without labels. 

\begin{figure}[ht]
    \centering
    \includegraphics[width=0.8\textwidth]{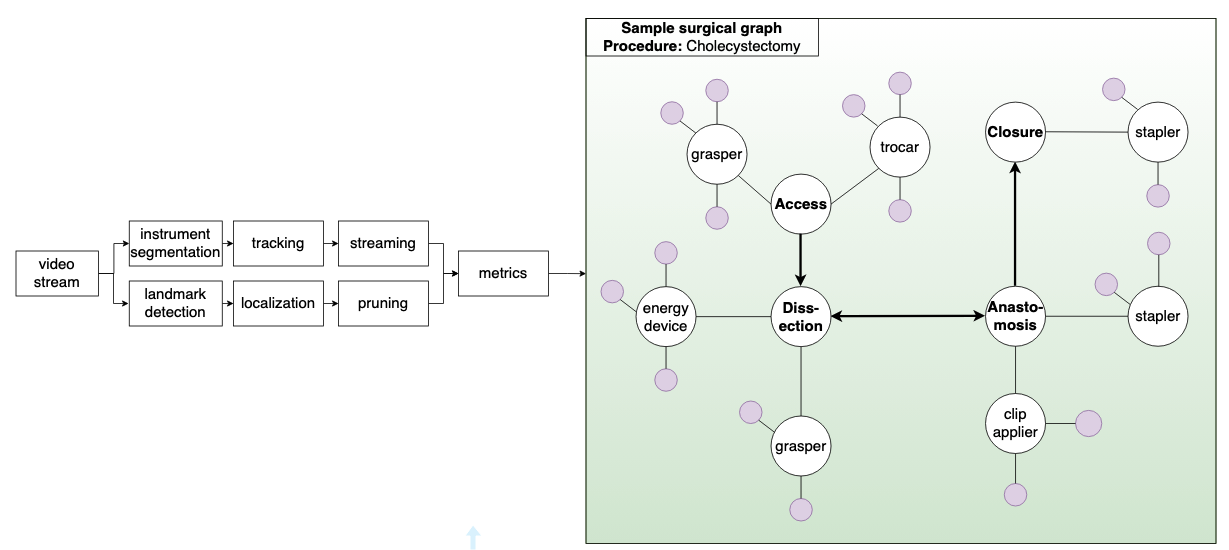}
    \caption{Our content extraction pipeline for defining the nodes of the proposed GNN architecture.}
    \label{fig:extract}
\end{figure}

\section{Experiments}
\subsection{Datasets}

\paragraph{EndoVis19}
The EndoVis19 dataset was created for endoscopic video analysis in surgical scenarios. The dataset consists of over 60 hours of endoscopic video recordings of 10 different surgical procedures, including cholecystectomy, appendectomy, and hernia repair. The videos were captured using various endoscopic cameras, and include both raw and annotated videos, as well as surgical tool and instrument segmentation masks. It is notable for its relatively large size and diverse set of surgical procedures, which enables the development and evaluation of algorithms for a wide range of endoscopic surgical scenarios. The dataset also includes expert annotations of surgical phases and actions, which can be used for developing and evaluating algorithms for automated surgical phase detection and action recognition.

\paragraph{Custom}
Our original dataset contains 309 sample clips, evenly distributed among the three major categories of \textit{novice}, \textit{intermediate}, and \textit{expert}. This dataset was procured as part of data-sharing agreements that [ANON] has with its partner institutions and is limited to laparoscopic cholecystectomy procedures. The videos are double-rated and annotated by skilled clinical analysts. 

\subsection{Pre-processing}
We direct the reader to Figure \ref{fig:cluster} and describe the steps to create a surgical graph from a surgical video:

\paragraph{Video segmentation}: Each surgical video needs to be segmented into different phases or steps, such as the initial incision, tissue dissection, and wound closure. This can be done manually or using automated techniques such as motion detection or deep learning algorithms. For the purposes of this paper, this was done manually and each procedure was broken down into two major components -- the calot and the dissection phases.

\paragraph{Graph construction}: Once the different surgical phases have been identified, they are represented as nodes in a graph, and the relationships between them are captured as edges. For example, an edge created between calot and dissection phases indicates that the tissue has been cut and is about to be dissected.

\begin{figure*}
\captionsetup[subfigure]{labelformat=empty}
\centering
\begin{subfigure}{.135\linewidth}
    \includegraphics[width=\linewidth]{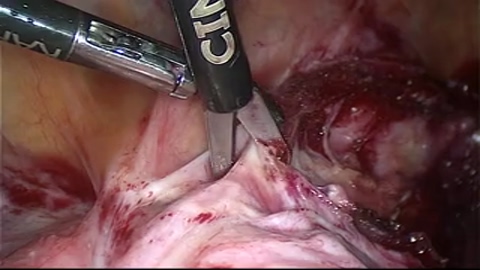}
\end{subfigure}
\vline
\begin{subfigure}{.135\linewidth}
    \includegraphics[width=\linewidth]{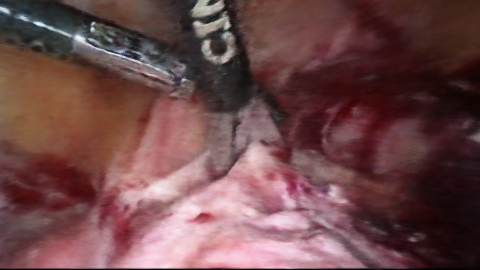}
\end{subfigure}
\begin{subfigure}{.135\linewidth}
    \includegraphics[width=\linewidth]{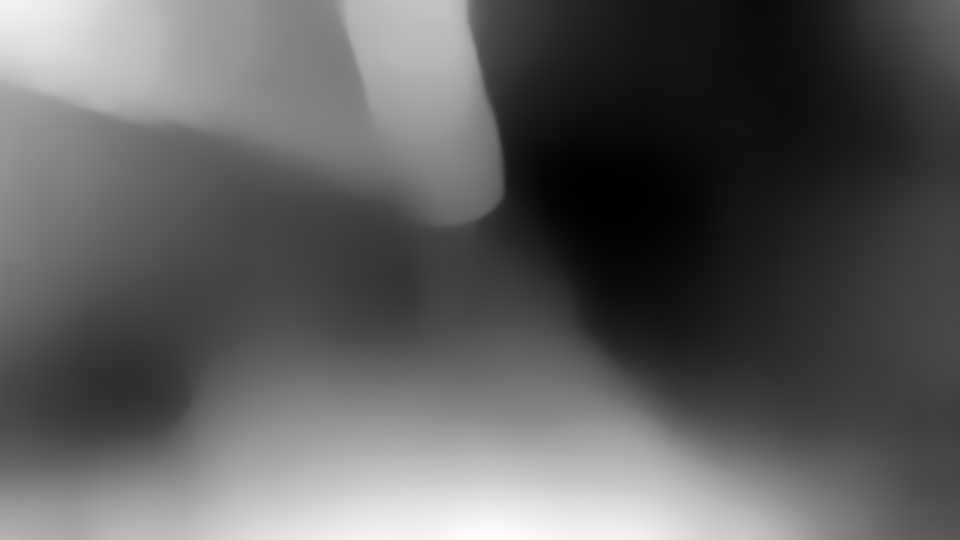}
\end{subfigure}
\begin{subfigure}{.135\linewidth}
    \includegraphics[width=\linewidth]{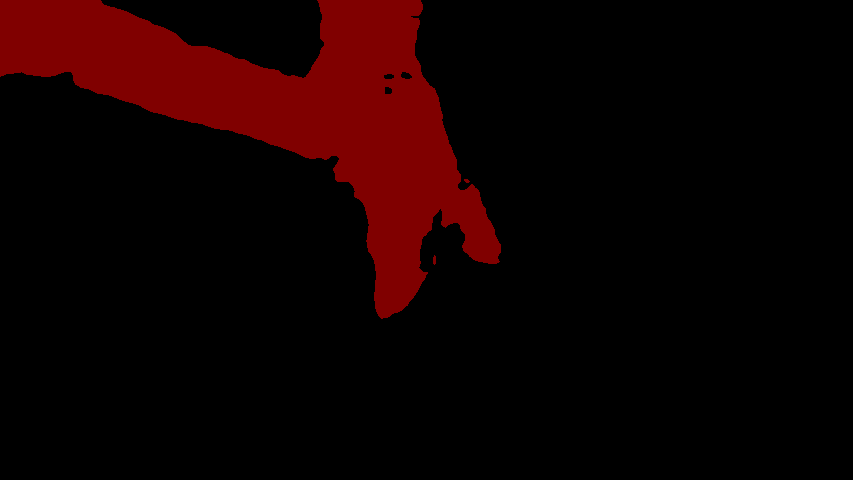}
\end{subfigure}
\vline
\begin{subfigure}{.135\linewidth}
    \includegraphics[width=\linewidth]{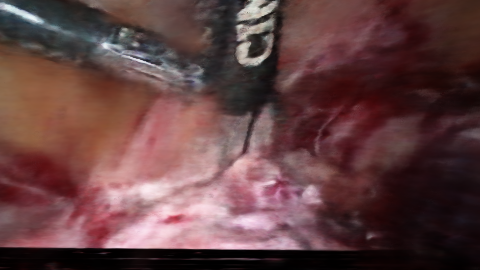}
\end{subfigure}
\begin{subfigure}{.135\linewidth}
    \includegraphics[width=\linewidth]{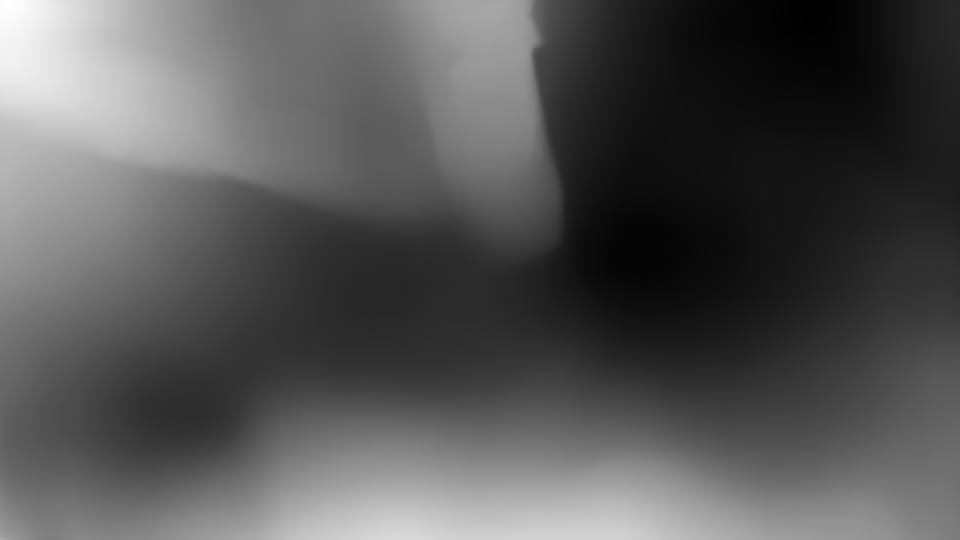}
\end{subfigure}
\begin{subfigure}{.135\linewidth}
    \includegraphics[width=\linewidth]{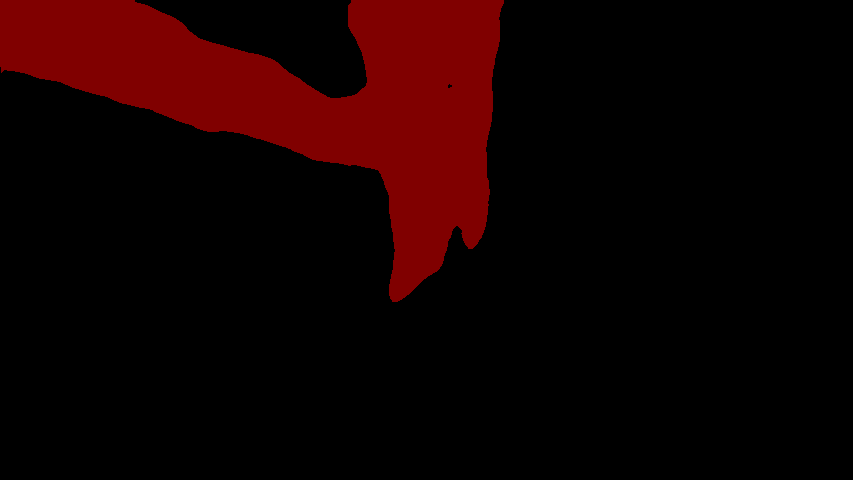}
\end{subfigure}

\begin{subfigure}{.135\linewidth}
    \includegraphics[width=\linewidth]{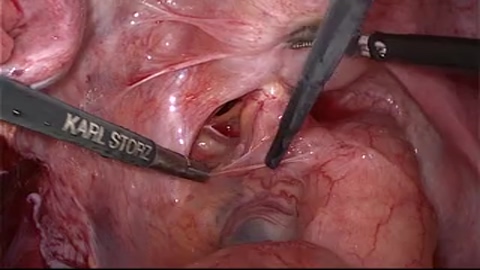}
\end{subfigure}
\vline
\begin{subfigure}{.135\linewidth}
    \includegraphics[width=\linewidth]{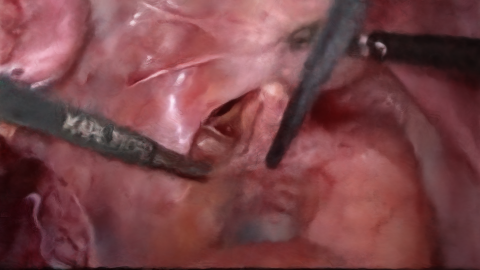}
\end{subfigure}
\begin{subfigure}{.135\linewidth}
    \includegraphics[width=\linewidth]{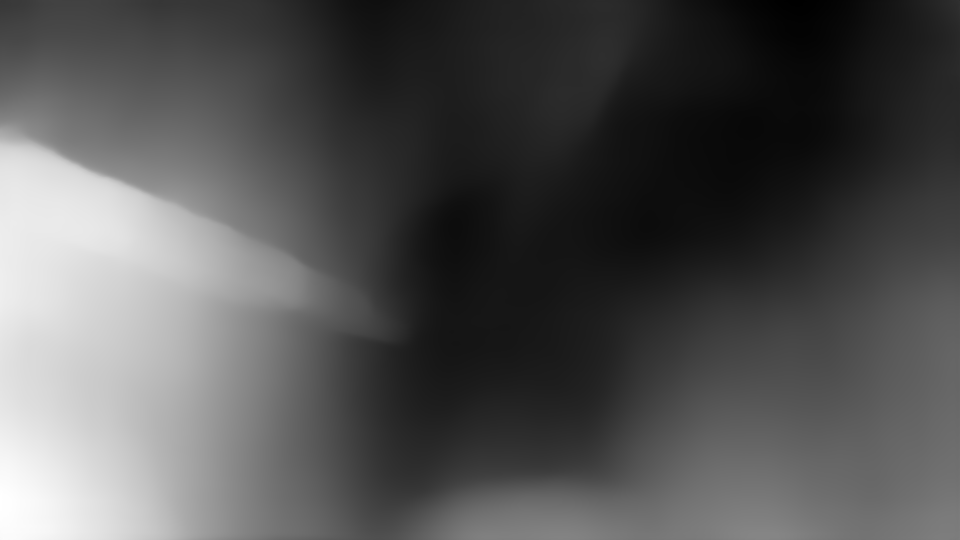}
\end{subfigure}
\begin{subfigure}{.135\linewidth}
    \includegraphics[width=\linewidth]{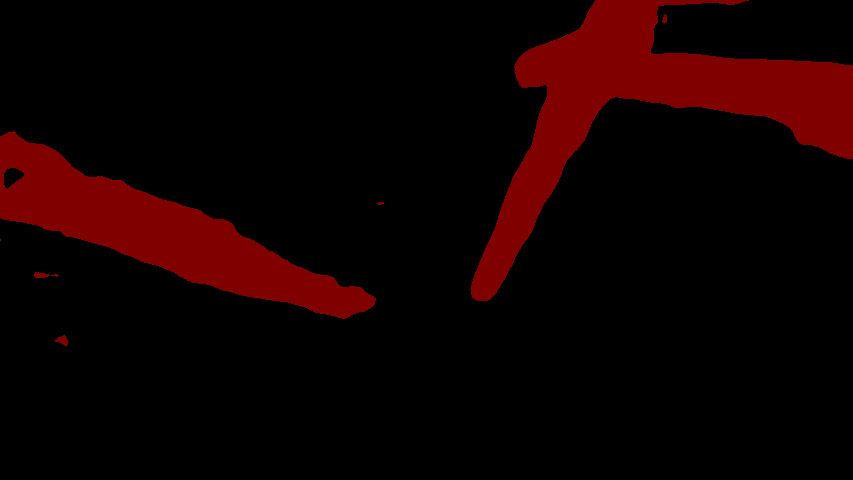}
\end{subfigure}
\vline
\begin{subfigure}{.135\linewidth}
    \includegraphics[width=\linewidth]{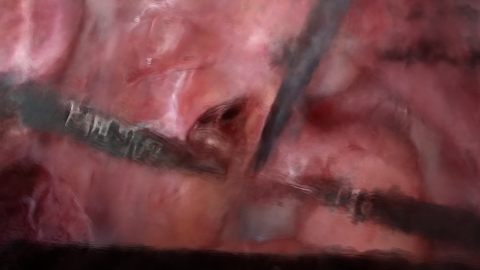}
\end{subfigure}
\begin{subfigure}{.135\linewidth}
    \includegraphics[width=\linewidth]{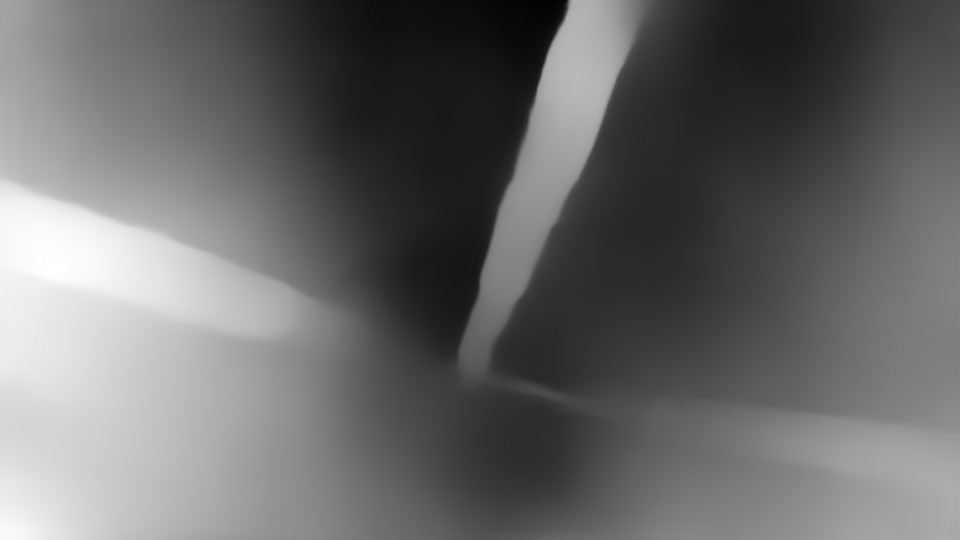}
\end{subfigure}
\begin{subfigure}{.135\linewidth}
    \includegraphics[width=\linewidth]{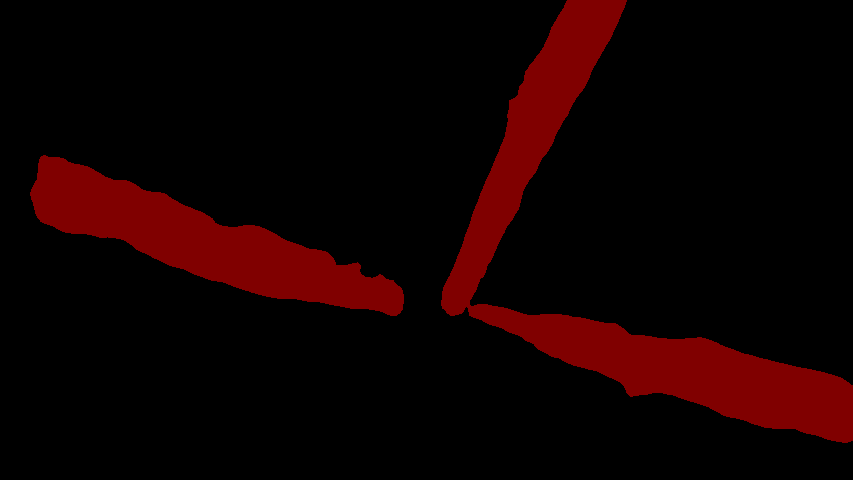}
\end{subfigure}

\begin{subfigure}{.135\linewidth}
    \includegraphics[width=\linewidth]{figures/sample1/gt.jpg}
    \caption{GT}
\end{subfigure}
\vline
\begin{subfigure}{.135\linewidth}
    \includegraphics[width=\linewidth]{figures/sample1/v000_t000.png}
    \caption{recon.}
\end{subfigure}
\begin{subfigure}{.135\linewidth}
    \includegraphics[width=\linewidth]{figures/sample1/00000.png}
    \caption{depth}
\end{subfigure}
\begin{subfigure}{.135\linewidth}
    \includegraphics[width=\linewidth]{figures/sample1/mask_00000.png}
    \caption{mask}
\end{subfigure}
\vline
\begin{subfigure}{.135\linewidth}
    \includegraphics[width=\linewidth]{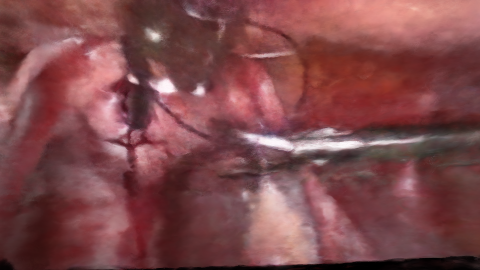}
    \caption{recon.}
\end{subfigure}
\begin{subfigure}{.135\linewidth}
    \includegraphics[width=\linewidth]{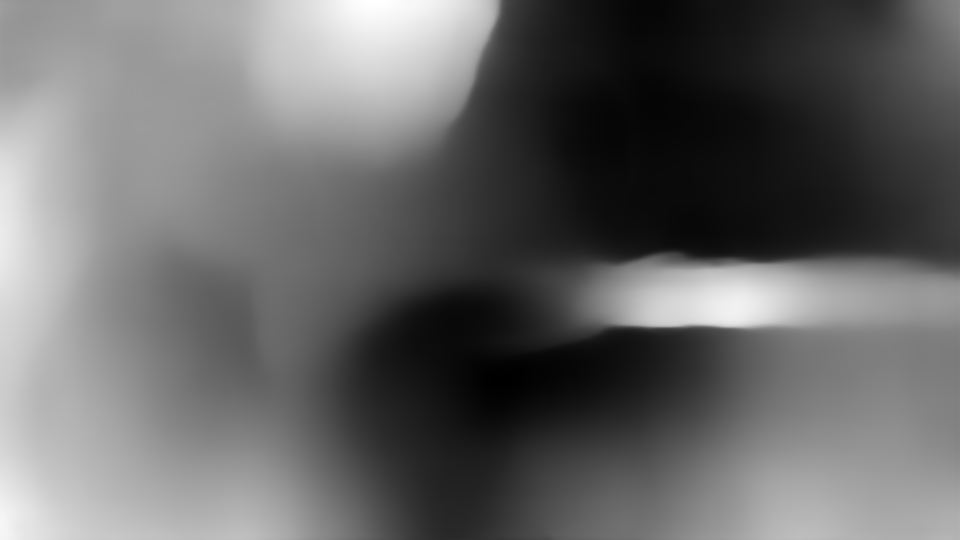}
    \caption{depth}
\end{subfigure}
\begin{subfigure}{.135\linewidth}
    \includegraphics[width=\linewidth]{figures/sample1/mask_00001.png}
    \caption{mask}
\end{subfigure}
\caption[short]{\textbf{3D representations}: We include the reconstructed images at times t=0 \textit{left} and t=1 \textit{right} for 3 cases, along with the corresponding instrument segmentation masks and depth masks. Note: Since this is a fixed camera view rendering, the ground truth image corresponds to the image at t=0.}
\label{fig:3d}
\end{figure*}

\paragraph{Feature extraction}: Various features can be extracted from the surgical graph, such as the frequency and duration of each action, the number of edges between different actions, and the overall structure of the graph. We extract this information directly by extracting instrument motion statistics across frames such as the position and speed of the instrument over time, illustrated in figure \ref{fig:extract}. For the purposes of this paper, we compare and contrast 2 approaches, namely segmentation assisted 2D \cite{khalid2022or} and radiance fields assisted 3D feature extraction \cite{gao2021dynamicnerf, pumarola2021dnerf, li2021neuralnsff, park2021hypernerf, khalid2022wildnerf, khalid2023refinerf} as described in Figure \ref{fig:3d}. For the 3D case, video clips were limited to 1 minute segments. This is because of the extremely computationally intensive nature of the reconstructions. Please refer to the Appendix for further details.

\paragraph{Data balancing}: The extracted features and their associated features are balanced by creating synthetic data using the ADASYN \cite{he2008adasyn} algorithm with seven $k$-neighbours.
\paragraph{Graph-based analysis}: The features extracted from the surgical graph can be used to assess various aspects of surgical skill, such as the speed and accuracy of the surgeon's movements, the level of coordination between different actions, and the complexity of the procedure.

\subsection{Training}

\paragraph{Supervised learning}
The architecture is shown in Figure \ref{fig:arch}. We use the Adam optimizer \cite{kingma2014adam} with an initial learning rate of 0.0025 which decays at 200 and 400 epochs with a batch size of 32. We train each model for a total of 1000 epochs on one Nvidia RTX 3090 card. 

\paragraph{Self-supervised learning} One approach to training graph neural networks (GNNs) is to use spectral self-supervised learning. This technique involves using the graph Laplacian matrix to generate embeddings for each node in the graph, which can then be used to train the GNN. 

To implement spectral self-supervised learning for GNNs, we can use the following steps:
\begin{itemize}
    \item Construct the adjacency matrix $A$ for the graph of interest.
    \item Compute the graph Laplacian matrix $L$ from the adjacency matrix using the formula $L = D - A$, where $D$ is the diagonal matrix of node degrees.
    \item Compute the eigenvectors and eigenvalues of the Laplacian matrix $L$.
    \item Select the $k$ smallest eigenvalues and corresponding eigenvectors to form the spectral embedding matrix $H$.
    \item Train the GNN using the spectral embedding matrix $H$ as illustrated in equation \ref{eq:spectral}.
\end{itemize}


where $\mathbf{X}_i$ and $\mathbf{X}j$ are the node features of nodes $i$ and $j$ in the graph, $w_{ij}$ is the weight of the edge connecting nodes $i$ and $j$, and $\mathbf{Z}_i$ and $\mathbf{Z}_j$ are the corresponding embeddings generated by the GNN. The spectral loss function aims to minimize the distance between the node embeddings in the learned feature space while taking into account the graph structure.

An important advantage of spectral self-supervised learning is that it can be used to learn embeddings that capture the global structure of the graph, rather than just local node features. This can be particularly useful in domains where the relationships between nodes are complex and non-local, as in the case of surgical videos.

\section{Explainability}
\begin{figure}[ht]
    \centering
    \includegraphics[width=0.8\textwidth]{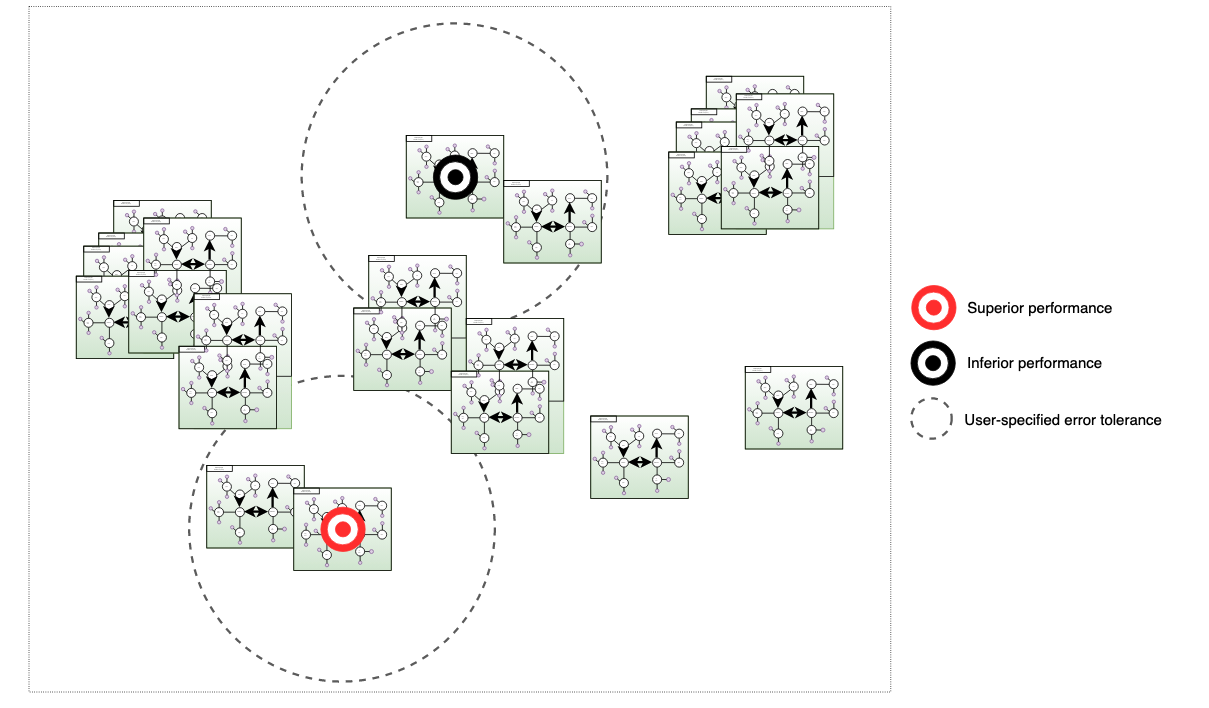}
    \caption{Explainable clusters in of surgical procedures allows for surgeries to be assessed and ranked.}
    \label{fig:cluster}
\end{figure}

Explainability is the ability of a model to provide understandable and transparent reasoning for its predictions. In the case of GNNs, explainability is particularly important because the structure of the input data is often essential to the prediction. Therefore, understanding how the model arrives at its prediction can provide valuable insights into the underlying data and the decision-making process. We visualize the learned node embeddings or feature representations, which are the intermediate outputs of SurGNN. By visualizing these embeddings, we gain insights into the relationships between nodes in the graph and the importance of different features for the prediction. We visualize the structure of the embeddings and project them in 2D space in Figure \ref{fig:cluster}. The assigned classifications are thus accompanied by supporting visualizations which can be invaluable for surgeons interested in improving their skills by comparing their performance to that of experienced surgeons. 

\section{Results and discussion}
We present our findings on two  datasets -- the custom dataset and the EndoVis19 dataset -- in Tables \ref{table:sst} and \ref{table:endovis} respectively. In both, each method is evaluated for different categories, and we report performance metrics such as Pearson correlation coefficients, Spearman rank correlation coefficients, Kendall's tau, precision, recall, and F1 scores. For each dataset, the baseline results are calculated by sampling from a Gaussian distribution and we average over 10 unique runs. To the best of our knowledge, there aren't any published baselines in literature that we can compare our approach against. We thus present the results of our supervised and self-supervised approaches.

\paragraph{[ANON] dataset} The baseline method performs poorly in terms of Spearman and Kendall correlation coefficients, with a mean value of $0.045$ for the overall score category. We include another approach, "SurGNN (SS)", which is a self-supervised approach. In contrast, our proposed method, represented by "SurGNN (S)" in the table, shows a significant improvement across all performance metrics and outperforms the baseline method in terms of precision, recall, and F1.

\paragraph{EndoVis19 dataset} The results show that our proposed method, represented by "SurGNN (S)," again outperforms  the baseline method across all performance metrics. For instance, for the Overall category, our proposed method achieves Pearson, Spearman, and Kendall's tau coefficients of 0.709, 0.711, and 0.678, respectively, while the baseline method has Spearman correlation of only $0.088$. Similarly, for the Time and Motion category, our proposed method outperforms the baseline method across all metrics. 

Overall, our results suggest that our proposed method,  "SurGNN (S)," is effective in improving performance on both datasets for most categories and performance metrics. However, the method's performance may vary depending on the category and the dataset. For example, the 2D analysis shows signifcantly better results than the 3D analysis. This could be a results of a few factors: The 3D analysis is limited to a length of 30s or approximately 60 frames in total (sampling 2 frames/s). The 30s segment chosen is also a potential source of error as it was chosen using clinical judgement. Longer samples were shown to lead to image degeneration, decreasing the quantity of available samples to 211 and 50 samples for the ANON and EndoVis19 datasets respectively.  



\begin{table}[ht]
\caption{Model performance on [ANON] dataset. \textit{SurGNN (S)} refers to our results using the supervised approach whereas \textit{SurGNN (SS)} refers to the proposed Self-supervised clustering approach detailed in the previous section.}
\resizebox{\textwidth}{!}{%
\centering
\begin{tabular}{llcccccccc}
\toprule
Method & Category & Mode & N & Pearson & Spearman & Kendall & Precision & Recall & F1 \\
\midrule 
- & Overall (baseline) & - & - & - & 0.045 & - & 0.140 & 0.140 & 0.140 \\
SurGNN (S) & Overall                          & 2D &  309 & 0.724 & 0.717 & 0.671 & 0.690 & 0.690 & 0.680 \\
SurGNN (S) & Precision of Operating technique & 2D &  309 & 0.731 & 0.724 & 0.675 & 0.670 & 0.680 & 0.675 \\
SurGNN (S) & Economy of movements             & 2D &  309 & 0.682 & 0.669 & 0.642 & 0.681 & 0.659 & 0.670 \\
SurGNN (SS) & Overall                         & 2D &  309 & 0.424 & 0.419 & 0.411 & 0.530 & 0.580 & 0.550 \\
SurGNN (SS) & Precision of Operating technique& 2D &  309 & 0.444 & 0.413 & 0.397 & 0.580 & 0.520 & 0.550 \\
SurGNN (SS) & Economy of movements            & 2D &  309 & 0.415 & 0.412 & 0.406 & 0.560 & 0.540 & 0.550 \\
SurGNN (S) & Overall                          & 3D &  211 & 0.355 & 0.325 & 0.316 & 0.430 & 0.440 & 0.435 \\
SurGNN (S) & Precision of Operating technique & 3D &  211 & 0.328 & 0.315 & 0.321 & 0.430 & 0.440 & 0.435 \\
SurGNN (S) & Economy of movements             & 3D &  211 & 0.409 & 0.401 & 0.389 & 0.410 & 0.410 & 0.410 \\
SurGNN (SS) & Overall                         & 3D &  211 & 0.288 & 0.273 & 0.279 & 0.320 & 0.320 & 0.320 \\
SurGNN (SS) & Precision of Operating technique& 3D &  211 & 0.261 & 0.254 & 0.255 & 0.300 & 0.310 & 0.300 \\
SurGNN (SS) & Economy of movements            & 3D &  211 & 0.214 & 0.197 & 0.196 & 0.310 & 0.310 & 0.310 \\
\bottomrule
\label{table:sst}
\end{tabular}
}
\end{table}

\begin{table}[ht]
\caption{Model performance on EndoVis19 dataset - \textit{SurGNN (S)} refers to our results using the supervised approach whereas \textit{SurGNN (SS)} refers to the proposed self-supervised clustering approach detailed in the previous section.}
\resizebox{\textwidth}{!}{%
\centering
\begin{tabular}{llcccccccc}
\toprule
 Method & Category & Mode & N & Pearson & Spearman & Kendall & Precision & Recall & F1 \\
\midrule 
 - & Overall (baseline) & - & - & - & 0.088 & - & 0.180 & 0.180 & 0.180 \\
 SurGNN (S) & Overall                 & 2D & 69 & 0.709 & 0.711 & 0.678 & 0.700 & 0.700 & 0.690 \\
 SurGNN (S) & Time and Motion         & 2D & 69 & 0.712 & 0.705 & 0.686 & 0.800 & 0.800 & 0.790 \\
 SurGNN (S) & Instrument Handling     & 2D & 69 & 0.663 & 0.677 & 0.628 & 0.660 & 0.670 & 0.650 \\
 SurGNN (SS) & Overall                & 2D & 69 & 0.409 & 0.394 & 0.381 & 0.500 & 0.520 & 0.510 \\
 SurGNN (SS) & Time and Motion        & 2D & 69 & 0.387 & 0.368 & 0.365 & 0.470 & 0.550 & 0.510 \\
 SurGNN (SS) & Instrument Handling    & 2D & 69 & 0.430 & 0.381 & 0.675 & 0.510 & 0.520 & 0.520 \\
 SurGNN (S) & Overall                 & 3D & 50 & 0.311 & 0.297 & 0.281 & 0.360 & 0.350 & 0.360 \\
 SurGNN (S) & Time and Motion         & 3D & 50 & 0.324 & 0.312 & 0.309 & 0.360 & 0.360 & 0.360 \\
 SurGNN (S) & Instrument Handling     & 3D & 50 & 0.317 & 0.316 & 0.297 & 0.320 & 0.320 & 0.320 \\
 SurGNN (SS) & Overall                & 3D & 50 & 0.156 & 0.144 & 0.121 & 0.250 & 0.250 & 0.250 \\
 SurGNN (SS) & Time and Motion        & 3D & 50 & 0.169 & 0.154 & 0.142 & 0.275 & 0.275 & 0.275 \\
 SurGNN (SS) & Instrument Handling    & 3D & 50 & 0.165 & 0.155 & 0.140 & 0.270 & 0.270 & 0.270 \\
\bottomrule
\label{table:endovis}
\end{tabular}
}
\end{table}

\subsection{Limitations and future directions}
We aim to provide a structured basis for research in visual scene understanding for surgical applications. Below are some items that should be considered before widespread application of these systems.

\paragraph{Inter-observer variability}: Different expert surgeons may have varying opinions on what constitutes good or bad surgical skill, leading to inconsistent ratings across different evaluators. Each dataset used in this analysis has been rated by multiple analysts and has thus gone through a rigorous quality assurance process.
\paragraph{Lack of granularity}: Subjective labels may provide only a high-level assessment of surgical performance, such as overall performance or proficiency in a particular task, without providing detailed feedback on specific actions or movements.
\paragraph{Bias}: Human evaluators may unconsciously or consciously introduce bias into their ratings, for example, by favoring or discriminating against certain individuals or groups based on factors such as gender, race, or ethnicity.
\paragraph{Limited feedback}: Subjective labels may not provide specific feedback on how to improve performance, which can be crucial for surgical trainees to learn and improve.
\paragraph{Time-consuming}: Collecting subjective ratings from expert evaluators can be time-consuming and expensive, especially when evaluating large datasets.
\paragraph{Concurrent validity}: The datasets presented in this paper consist primarily of cholecystectomy procedures. Testing the efficacy of our proposed method on datasets with different types of demographics is an important line of future work.  

To address these issues, objective and quantitative measures of surgical performance, such as motion analysis or tool usage, have been proposed. These measures can provide more granular and unbiased assessments of surgical skill, and can be automated using computer vision and machine learning techniques. Additionally, such objective measures can provide specific feedback on how to improve performance and can facilitate more efficient evaluation of large datasets.

\section{Conclusion}
We present two distinct approaches using graph neural networks (GNNs) for explainable visual scene understanding in surgical skill assessment. We demonstrate that, by representing surgical procedures as graph structures and training them using supervised and self-supervised techniques, relevant features can be extracted from the complex visual data and used to predict various aspects of surgical skill. GNNs also allow for interpretable results by identifying the specific actions, instruments, or anatomical structures that contribute to the predicted skill metrics. We highlight the limitations of traditional assessment methods and how GNN-based models can provide a more objective and fine-grained analysis of surgical performance. Explainable visual scene understanding models for surgical skill assessment represents an exciting opportunity to leverage cutting-edge AI techniques to improve surgical education, training, and outcomes.



\clearpage


{\small
\bibliographystyle{ieee_fullname}
\bibliography{neurips_2022}
}

\end{document}